\pdfoutput=1

\documentclass[runningheads]{llncs}
\usepackage{times}
\usepackage{graphicx}
\usepackage{amssymb} 
\usepackage{color}
\usepackage{multirow}
\usepackage{grffile}
\usepackage{placeins}

\usepackage{array}
\usepackage{rotating}
\usepackage[latin1]{inputenc}
\usepackage{color}

\usepackage[width=122mm,left=12mm,paperwidth=146mm,height=193mm,top=12mm,paperheight=217mm]{geometry}

\begin{document}
\def\x{{\mathbf x}}
\def\L{{\cal L}}

\newcolumntype{L}[1]{>{\raggedright\let\newline\\\arraybackslash\hspace{0pt}}m{#1}}
\newcolumntype{C}[1]{>{\centering\let\newline\\\arraybackslash\hspace{0pt}}m{#1}}
\newcolumntype{R}[1]{>{\raggedleft\let\newline\\\arraybackslash\hspace{0pt}}m{#1}}

\newcommand{\TODO}[1]{\textcolor{red}{TODO:#1}}
\newcommand{\NEW}[1]{\textcolor{blue}{NEW:#1}}
\newcommand{\HIGH}[1]{\textbf{\textcolor{red}{--NOTE: --}}\textcolor{red}{#1}}

\newcommand{\fig}[1]{Figure~\ref{#1}}
\newcommand{\tab}[1]{Table~\ref{#1}}
\newcommand{\subfig}[2]{Fig.~\ref{#1}#2}
\newcommand{\subtab}[2]{Table~\ref{#1}#2}
\newcommand{\points}{...,}
\newcommand{\sect}[1]{Sect.~\ref{#1}}
\newcommand{\eq}[1]{Eq. (\ref{#1})}
\newcommand{\iii}{{\cal I}}
\newcommand{\bfp}{{\bf p}}
\newcommand{\etal}{\emph{et al. }} 
\newcommand{\ie}{i.e.} 
\newcommand{\eg}{e.g.} 
\newcommand{\Algorithm}[1]{Algorithm~\ref{#1}}
\newcommand{\mys}[1]{{\footnotesize{#1}}}

\DeclareGraphicsExtensions{.pdf}
\graphicspath{{./figs/}}

\hyphenation{DecomposeMe}
\hyphenation{AlexNetOWTBn}
\pagestyle{headings}
\mainmatter
\def\ECCV16SubNumber{686}  

\title{DecomposeMe: Simplifying ConvNets for End-to-End Learning} 

\titlerunning{DecomposeMe}

\authorrunning{Alvarez \& Petersson}

\author{Jose M. Alvarez and Lars Petersson}
\institute{NICTA / Data61, Canberra, Australia\\ \{jose.alvarez,lars.petersson\}@data61.csiro.au}
\maketitle
\vspace{-0.4cm}
\begin{abstract}
Deep learning and convolutional neural networks (ConvNets) have been successfully applied to most relevant tasks in the computer vision community. However, these networks are computationally demanding and not suitable for embedded devices where memory and time consumption are relevant.

In this paper, we propose {\textbf{DecomposeMe}}, a simple but effective technique to learn features using 1D convolutions. The proposed architecture enables both simplicity and filter sharing leading to increased learning capacity. A comprehensive set of large-scale experiments on ImageNet and Places2 demonstrates the ability of our method to improve performance while significantly reducing the number of parameters required. Notably, on Places2, we obtain an improvement in relative top-1 classification accuracy of $7.7\%$ with an architecture that requires 92\% fewer parameters compared to VGG-B. The proposed network is also demonstrated to generalize to other tasks by converting existing networks.

\end{abstract}


\vspace{-0.5cm}
\section{Introduction}
Deep Architectures and, in particular, convolutional neural networks (ConvNets) have experienced great success in recent years. However, while being able to successfully tackle a wide range of challenging problems, current architectures are often limited by the need for large amounts of memory and computational capacity.

In this paper, we set out to alleviate these issues where possible and, in particular, we consider ConvNets used for computer vision tasks, as typically the issues of memory and computation become paramount in that context. Networks useful for real-world tasks may sometimes require as much as a few hundred million parameters~\cite{sermanet-iclr-14} to produce state-of-the-art results which increases the memory footprint as well as the computational need. Unfortunately, this means that it is hard to deploy applications where memory and computational resources are relevant such as portable devices. In this work, we demonstrate that the use of filter compositions can not only reduce the number of parameters required to train large scale networks, but also provide better classification performance as evidenced by our experimental results.

We identify two bottlenecks in current convolutional neural network models: computation and memory. While the most computationally expensive operations occur in the first few convolutional layers~\cite{Fergus:NIPS2014}, the larger memory footprint is typically caused by the later, fully-connected, layers. Here, we focus on the first bottleneck mentioned and propose a new architecture intended to speed up the first set of convolutional layers while maintaining or surpassing the original performance. Further, as a consequence of the improved learning capacity of the network, our approach indirectly alleviates the second bottleneck leading to a significant reduction in the memory footprint.
\begin{figure}[!t]
\centering
\includegraphics[width=0.97\columnwidth]{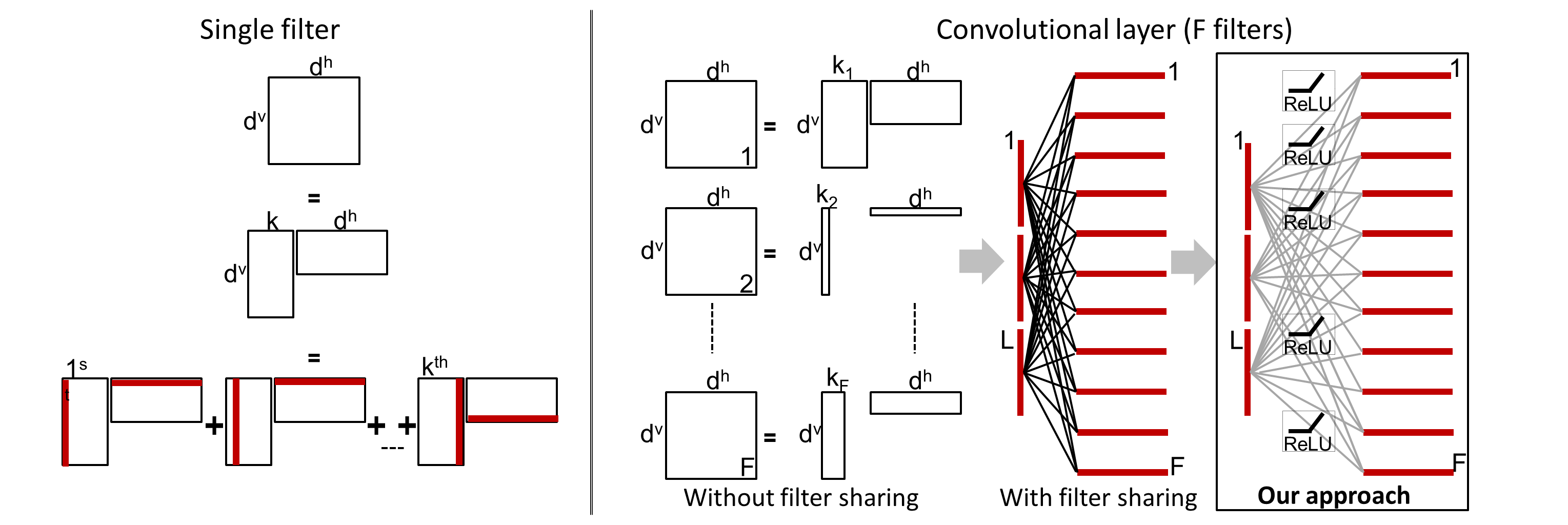}
\vspace{-0.5cm}\caption{\textbf{Left:} A 2D tensor (filter) of rank-k can be represented as the product of two matrices or, alternatively, as the linear combination of the outer product of a set of vectors. \textbf{Right:} A convolutional layer in a ConvNet consisting of $F$ rank-k  filters (not necessarily the same rank or rank-1) can be represented as a linear combination of 1D filters. Furthermore, these filters can be shared within the layer to minimize redundancy. Our approach benefits from this decomposition and, considers compositions of linearly rectified 1D filters.}
\label{fig:Diagram}\vspace{-0.5cm}
\end{figure}

Many of the current approaches attempting to reduce the computational need have relied on the hope that learned ND filters are low rank enough such that they can be approximated by separable filters~\cite{Fergus:NIPS2014,SepFilters:PAMI2015,Jaderberg14b}. The main advantages of these approaches are computational cost if the filters are large and reduction in the number of parameters in convolutional layers. However, these methods require a pre-trained network using complete filters and a post processing step to fine-tuning the network and minimize the drop in performance compared to the pre-trained network.

Our approach is different from those mentioned above. We propose {\textbf{DecomposeMe}}, a novel architecture based on 1D convolutions depicted in~\fig{fig:Diagram}. This architecture introduces three main novelties. \emph{i)} Our architecture relies on imposing separability as a hard constraint by directly learning 1D filter compositions. The fundamental idea behind our method is the fact that any matrix (2D tensor) can be represented as a weighted combination of separable matrices. Therefore, existing architectures can be adequately represented by composing 2D filter kernels by a combination of 1D filters (1D tensors). \emph{ii)} Our proposal further improves the compactness of the model by sharing filters within the convolutional layers.  In this way, the proposed network minimizes redundancy and thus further reduces the number of parameters. \emph{iii)} Our proposal improves the learning capacity of the model by inserting a non-linearity in between the 1D filter components. With this modification, the effective depth of the network increases which is intimately related to the number of linear regions available to approximate the sought after function~\cite{LinearRegionsNIPS2014}. As a result, we obtain compact models that do not require a pre-trained network and minimize the computational cost and the memory footprint compared to their equivalent networks using 2D filters. Reduced memory footprint has the additional advantage of enabling larger batch sizes at train time and, therefore, computing better gradient approximations leading to, as demonstrated in our experiments, improved classification performance.

A comprehensive set of experiments on four datasets including two large-scale datasets such as Places2 and ImageNet shows the capabilities of our proposal. For instance, on Places 2, compared to a VGG-B model, we obtain a relative improvement in top-1 classification accuracy of $7.7\%$ using $92\%$ fewer parameters compared to the baseline and with a speed up factor of $4.3x$ in a forward-backward pass. Additional experiments on stereo matching also demonstrate the general applicability of the proposed architecture.

\section{Related work}
\label{sect:sota}
\vspace{-0.25cm}
In the last few years, the computer vision community has experienced the great success of deep learning. The performance of these end-to-end architectures has continuously increased and outperformed traditional hand-crafted systems. An essential component to their success has been the increment of data available as well as the availability of more powerful computers making possible the training of larger and more computationally demanding networks. For instance, in 2012 the AlexNet~\cite{Krizhevsky_imagenetclassification} model was proposed and won the ImageNet classification challenge with a network that had approximately 2.0M parameters in the convolutional layers (i.e., excluding fully connected ones). More recently, different variations of VGG models were introduced~\cite{Simonyan14c} of which VGG-16 has over 14.5M feature parameters. VGG-16 increases the depth of the model by substituting each convolutional kernel with consecutive convolutions consisting of smaller kernels while maintaining the number of filters. As an example, the VGG models as in~\cite{Simonyan14c} substitutes $7\times 7$ kernels with $3$ consecutive rectified layers of $3\times 3$ kernels. This operation reduces the degrees of freedom compared to the original kernels but at the same time inserts a non-linearity in-between the smaller $3\times 3$ kernels increasing the capacity of the model for partitioning the space~\cite{LinearRegionsNIPS2014}. Despite improving the classification performance, the large number of parameters not only makes the training process slow but also makes it difficult to use these models in portable devices where memory and computational resources are relevant.

The growing number of applications deployed in portable devices has motivated recent efforts in speeding up deep models by reducing their complexity. A forerunner work on reducing the complexity of a neural network is the so-called network distillation method proposed in~\cite{distNets}. The idea behind this approach is to train a large, capable, but slow network and then refine this by taking the output of that to train a smaller one. The main strength comes from using the vast network to take care of the regularization process facilitating subsequent training operations. However, this method requires a large pre-trained network to begin with which is not always feasible especially in new problem domains.

Memory-wise the largest contribution comes from the fully connected layers while time-wise the bottleneck is in the first convolutional layers due to a large number of multiplications (larger kernels). In this work, we address the former by simplifying the first convolutional layers. There have been several attempts to reduce the computational cost of these first layers, for example, Denil et. al.~\cite{DenilSDRF13} proposed to learn only 5\% of the parameters and predict the rest based on dictionaries. The existence of this redundancy in the network has motivated other researchers to explore linear structures within the convolutional layers~\cite{Fergus:NIPS2014,JaderbergVZ14,Liu:2015} usually focusing on finding approximations to filters (low-rank filters) by adding constraints in a post-learning process. More specifically, these approaches often learn the unconstrained filter and then approximate the output using a low-rank constraint. For instance, ~\cite{Fergus:NIPS2014} and \cite{JaderbergVZ14} focus on improving test time by representing convolutional layers as linear combinations of a certain basis. As a result, at test time, a lower number of convolutions is needed to achieve some speeds ups with virtually no drop in performance. Liu et al~\cite{Liu:2015} instead consider sparse representations of the basis rather than linear combinations. However, similar to the distillation process, the starting point of these methods is a pre-trained model.

Directly related to our proposed method is~\cite{Rigamonti13a} although it is not a convolutional neural network. In that paper, the authors aim at learning separable filters for image processing. To this end, they propose learning a filter combination reinforcing filter separability using low-rank constraints in the cost function. Their results are promising and demonstrate the benefits of learning combinations of separable filters. In contrast to that work, we work within the convolutional layers of a neural network and our filter sharing strategy is different. More importantly, we do not use soft constraints during the optimization process. Instead, we directly enforce filters to be 1D.
\vspace{-0.3cm}\section{Simplifying ConvNets through Filter Compositions}
\label{sect:method}\vspace{-0.3cm}
In this section we present our DecomposeMe architecture. The essence of our proposal consists of decomposing the ND kernels of a traditional network into N consecutive layers of 1D kernels, see~\fig{fig:Diagram}. We consider each ND filter as a linear combination of other filters. In contrast to~\cite{Rigamonti13a} where they seek to find these other filters by solving an optimization problem with additional low-rank constraints, we impose the filters to be 1D and learn them directly from the data. Performance-wise, it turns out that such a decomposition not only mimics the behavior of the original, more complex, network but often surpasses it while being significantly more compact and experiencing a lower computational cost.

For the purpose of clarity, we will here consider 2D filters, however, the analysis is similarly applicable to the ND case. With that in mind, a typical convolutional layer can be analyzed as follows. Let $\mathbf{W} \in \mathbb{R}^{C \times d^h\times d^v\times F}$ denote the weights of a 2D convolutional layer where $C$ is the number of input planes, $F$ is the number of output planes (target number of feature maps) and $d^v\times d^h$ represent the kernel size of each feature map (usually $d^h=d^v\equiv d$). Let $b \in \mathbb{R}^F$ be the vector representing the bias term for each filter. Further, let us now denote $\mathbf{f}^i \in \mathbb{R}^{d^v\times d^h}$ as the ith kernel in the layer. Common approaches first learn these filters from data and then find low-rank approximations as a postprocessing step~\cite{JaderbergVZ14}. However, learned filters may not be separable \eg, specially those in the first convolutional layer~\cite{Fergus:NIPS2014,JaderbergVZ14}, and these algorithms require an additional fine tuning step to compensate drops in performance.

Instead, it is possible to relax the rank-1 constraint and essentially rewrite $\mathbf{f}^i$ as a linear combination of $1$D filters~\cite{Rigamonti13a}:
\begin{equation}
\mathbf{f}^i = \sum_{k=1}^K \sigma_k^i\bar{v}_k^i(\bar{h}_k^i)^T
\end{equation}
\noindent where $\bar{v}^i_k$ and $(\bar{h}^i_k)^T$ are vectors of length $d$, $\sigma^i_k$ is a scalar weight, and $K$ is the rank of $\mathbf{f}^i$.

Based on this representation we propose DecomposeMe which is an architecture consisting of decomposed layers. Each decomposed layer represents a N-D convolutional layer as a composition of 1D filters and, in addition, by including a non-linearity $\varphi(\cdot)$ in-between (\fig{fig:Diagram}). The i-th output of a decomposed layer, $a_i^1$, as a function of its input, $a_*^0$, can be expressed as:
\begin{equation}
\label{eq:dec}
a_i^1=\varphi(b_i^h + \sum_{l=1}^L{\bar{h}_{il}^T* [ \varphi( b_l^v + \sum_{c=1}^C{\bar{v}_{lc}*a_c^0})} ] )
\end{equation}
\noindent where $L$ represents the number of filters in the intermediate layer. $\varphi(\cdot)$ is set to rectified linear unit (ReLU~\cite{Krizhevsky_imagenetclassification}) in our experiments. 

Decomposed layers have two major properties, intrinsically low computational cost, and simplicity. \textbf{Computational cost:} Decomposed layers are represented with a reduced number of parameters compared to their original counterparts. This is an immediate consequence of two important concepts: the direct use of 1D filters and the sharing scheme across a convolutional layer leading to greater computational cost savings, especially for large kernel sizes. \textbf{Simplicity:} Decomposed architectures are deeper but simpler structures. Decomposed layers are based on filter compositions and therefore lead to smoother (simpler) 2D equivalent filters that help during training by acting as a regularizing factor~\cite{Fergus:NIPS2014}. Moreover, decomposed layers include a non-linearity in-between convolutions increasing the effective depth of the model. As a direct consequence, the upper bound of the number of linear regions available is increased~\cite{LinearStructs:NIPS2014}. Evident from our results, decomposed layers learn faster, as in per epoch, than equivalent 2D convolutional layers. This suggests that the simplicity of the decomposed layers not only reduces the number of parameters but also benefits the training process. 

Converting existing structures to decomposed ones is a straight forward process as each existing ND convolutional layer can systematically be decomposed into sets of consecutive layers consisting of $1$D linearly rectified kernels and $1$D transposed kernels as shown in \fig{fig:Diagram}. In the next section we apply decompositions to two well-known computer vision problems such as image classification and stereo matching.
\subsection{Complexity Analysis}\vspace{-0.1cm}
We analyze the theoretical speed up of the proposed method as follows: Consider as the baseline a convolutional layer of dimensions $C\times F$ with filters of spatial size of $d^v\times d^h$.
Without loss of generality, we can assume $d^h=d^v=d$. This baseline is then decomposed into two consecutive layers $C\times L$ and $L\times F$ with filter size $d\times 1$ and $1\times d$
respectively. The computational cost of these two schemes is proportional to $CFd^2$ and $L(C+F)d$ respectively. Therefore, considerable improvements are achieved
when $L(C+F)<<CFd$. The analysis of this expression reveals that, although, $d$ is larger in the first layer (\eg, $11$ for AlexNet~\cite{Krizhevsky14}), $C$ is usually too small compared to $L$ to make a
significant difference (\eg, $3$ for RGB images). Current architectures tend to have a large number of filters in later layers. For instance, consider a VGG model using kernels of size $3\times 3$, consecutive layers of equal size (\eg, $256$), and maintaining the number of output filters through the decomposed layer ($L=256$). In that case, the theoretical improvement of our method is given by $256(256 + 256)$ vs. $256\times 256 \times 3$.

\begin{figure}[!t]
\begin{center}
\begin{tabular}{ccc}
\hspace{-0.25cm}\includegraphics[width=0.32\columnwidth]{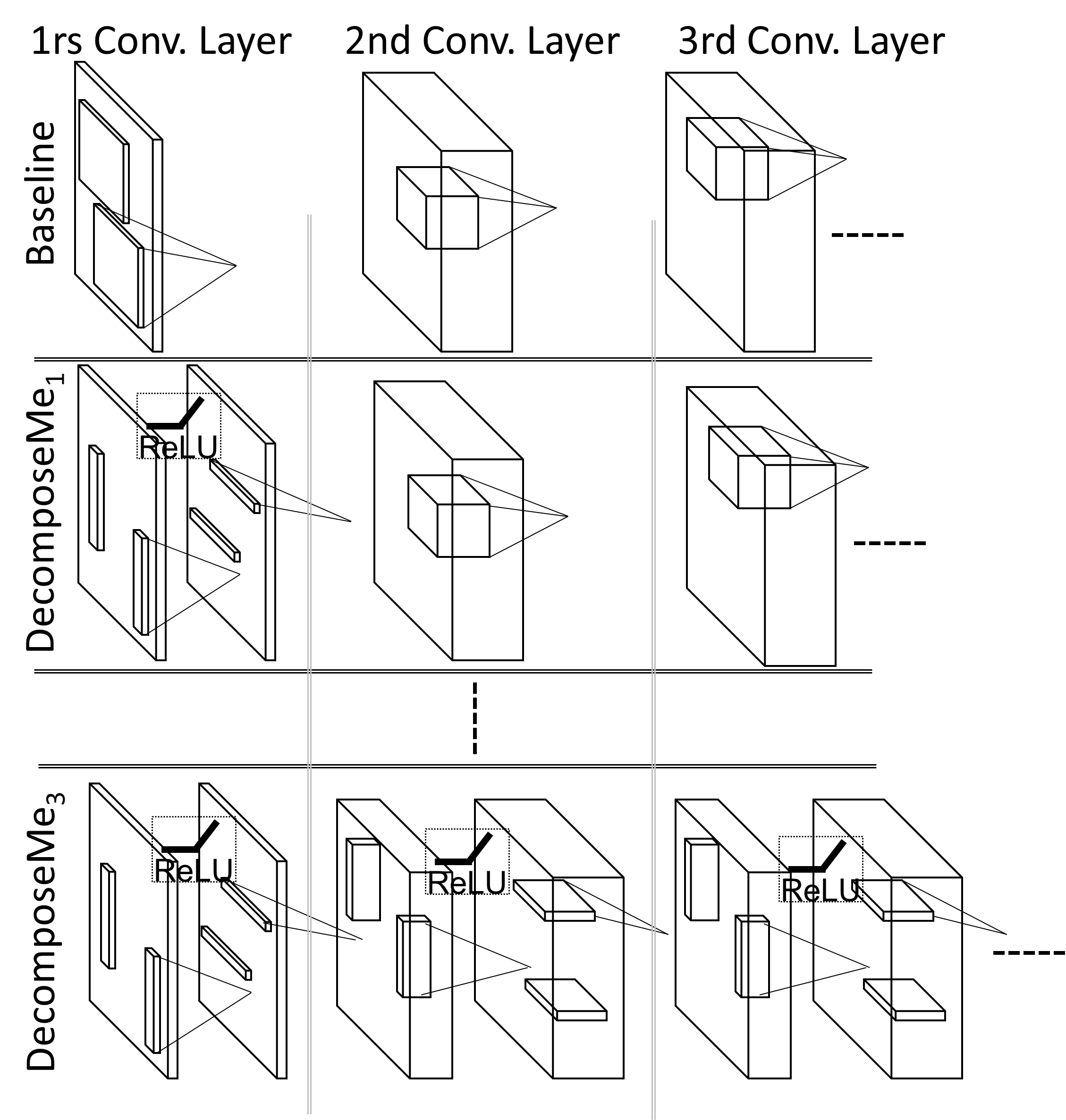}&
\hspace{1.05cm}\includegraphics[width=0.1\columnwidth]{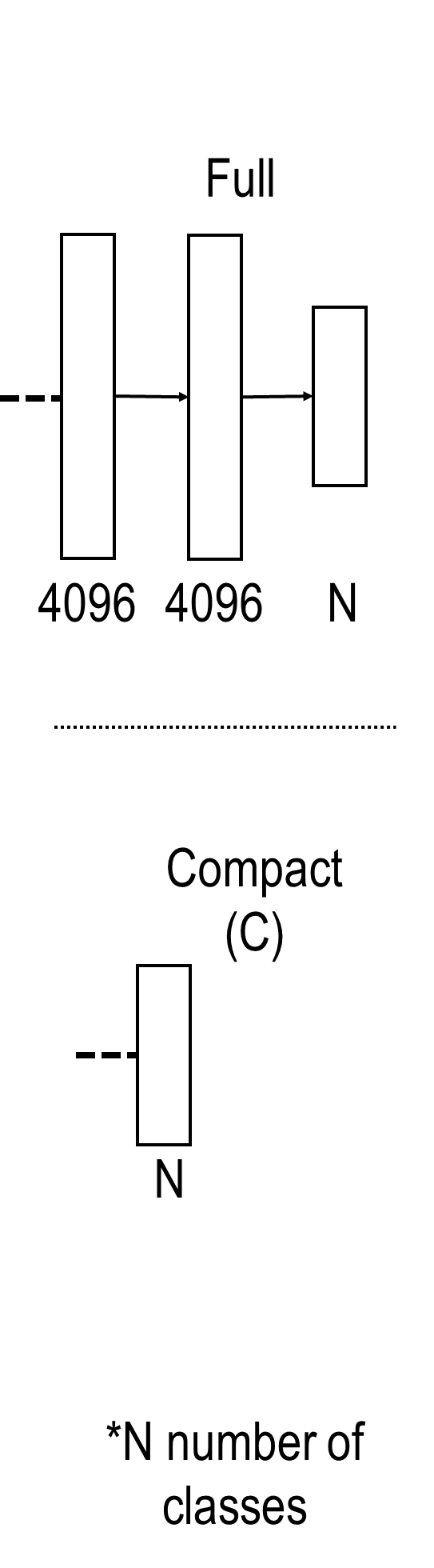}&\hspace{1.25cm}\includegraphics[width=0.33\columnwidth]{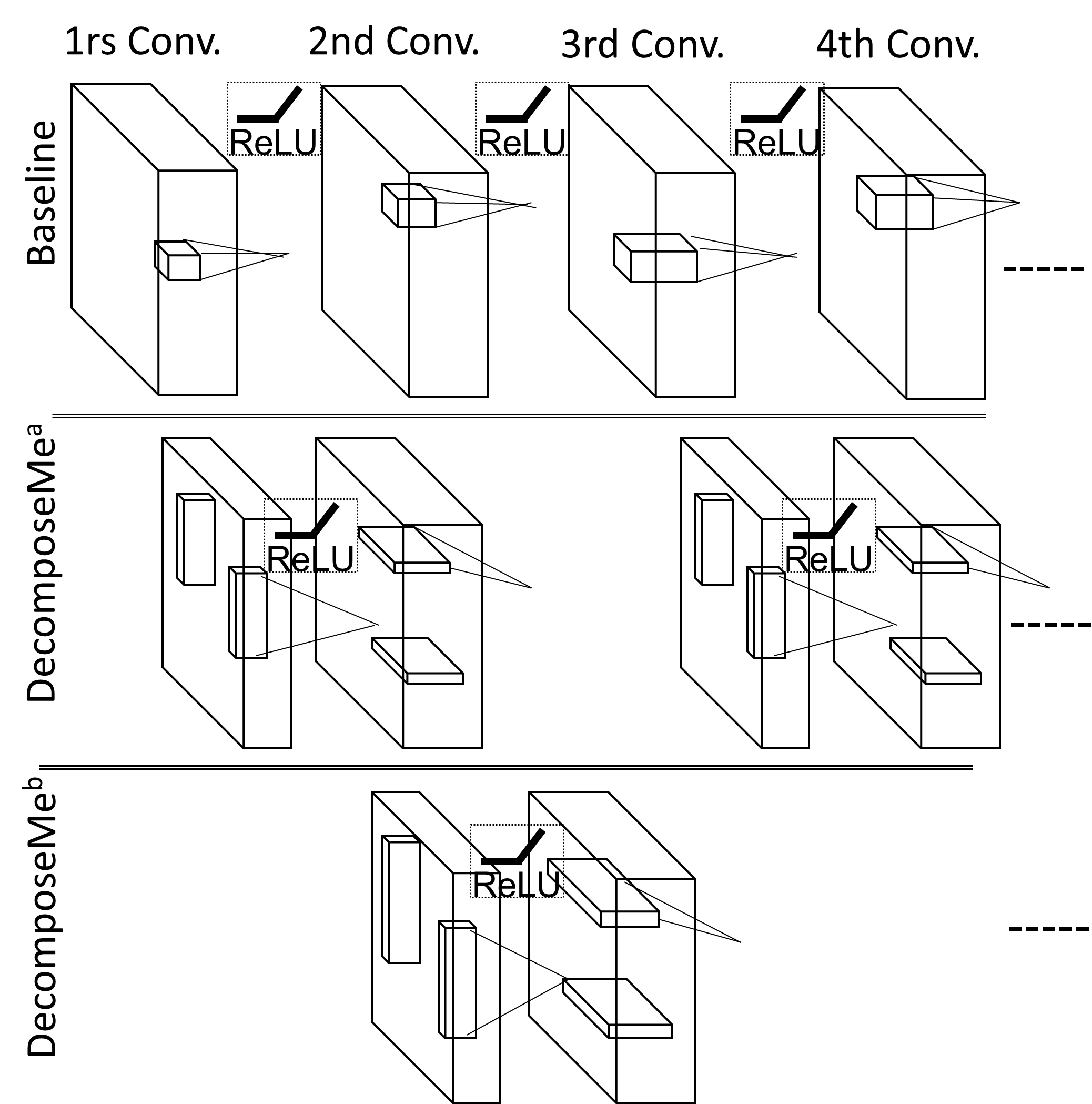}
\\
(a)&\hspace{1.25cm}(b)&\hspace{1.25cm}(c)\\
\end{tabular}
\end{center}
\vspace{-0.5cm}
\caption{\textbf{DecomposeME:} a) Convolutional layers in a ConvNet can be represented using our proposed DecomposeMe$_i$ architecture, where $i$ is the number of layers being decomposed. b) We evaluate the proposal using Full and Compact models. The former is the original architecture consisting of two fully connected layers and the final classification layer with $4096$,$4096$,$N$ neurons respectively; The latter, a Compact model, directly connects the output of the last convolutional layer to a layer with $N$ neurons. $N$ is the number of classes. c) Consecutive convolutional layers can also be converted into decomposed ones maintaining the size of the receptive field in the input feature map.}\label{fig:DecMEInstances}\vspace{-0.4cm}
\end{figure}
\section{Experiments}
\label{sect:experiments}
\vspace{-0.3cm}
We conduct two sets of experiments representing different use cases to validate our proposal. Firstly, we run experiments performing image classification. More specifically, we test four well-known network architectures, namely LeNet~\cite{LeNet}, CIFAR-10 quick~\cite{quick} and AlexNet~\cite{Krizhevsky_imagenetclassification} and VGG~\cite{Simonyan14c}, on three publicly available datasets; MNIST~\cite{MNIST}, CIFAR-10~\cite{CIFAR-10} and ImageNet~\cite{ImageNet}. An additional experiment is included on the challenging Places2 dataset~\cite{Places2}. Secondly, we run experiments performing stereo matching to show the generic learning capabilities and applicability of our proposal. To this end, we consider a state-of-the-art stereo matching problem and replace the existing, accurate, network of~\cite{ZbontarL15} with our decomposed architecture. This set of experiments is carried out on the KITTI benchmark~\cite{Geiger2012CVPR}.

\subsection{Image Classification}
\vspace{-0.3cm}
All the experiments on image classification are conducted on a Dual Xeon 8-core E5-2650 with 128GB of RAM using two Kepler Tesla K20 GPUs in parallel, unless otherwise specified. We use the torch-7 framework~\cite{Collobert_NIPSWORKSHOP_2011} and large-scale experiments are carried out using the multi-GPU implementation available in~\cite{ImageNetCode}. Learning rate, weight decay and momentum were set to the default values. More precisely, we start with a learning rate of $0.01$ which is decreased when the training error plateaus; weight decay is set to $0.0001$ and momentum to $0.9$. Again, unless otherwise specified, we use the same hyper-parameter setup as in the original experiments. Data augmentation is done through random crops where necessary and random horizontal flips with probability $0.5$. Please note that other training approaches may use different data augmentation techniques such as color augmentation~\cite{Krizhevsky_imagenetclassification}. For a fair comparison, we select the original networks as baselines and all models including baselines are trained from scratch on the same computer using the same seed and the same framework.

A basic decomposed layer consists of vertical kernels followed by horizontal ones, and non-linearities in-between 1D convolutions are set to rectifier linear units (ReLU). We evaluate different instances of this model referred to as DecomposeMe$_i^k$ where the sub-index is the number of layers being decomposed (\fig{fig:DecMEInstances}a), and the super-index indicates variations in the composition of the layer such as kernel size, the non-linearity being used or the order of the kernels. Decompositions respect the size of the filter in the original model, and the number of output filters from the convolutional layer is maintained. Layers that are not decomposed are left as in the original model. For specific experiments we show results for variations within each of these instances.

\begin{figure}[!t]
\centering
\begin{tabular}{cc}
\begin{minipage}[c]{0.50\textwidth}
\hspace{-0.15cm}\includegraphics[width=\columnwidth]{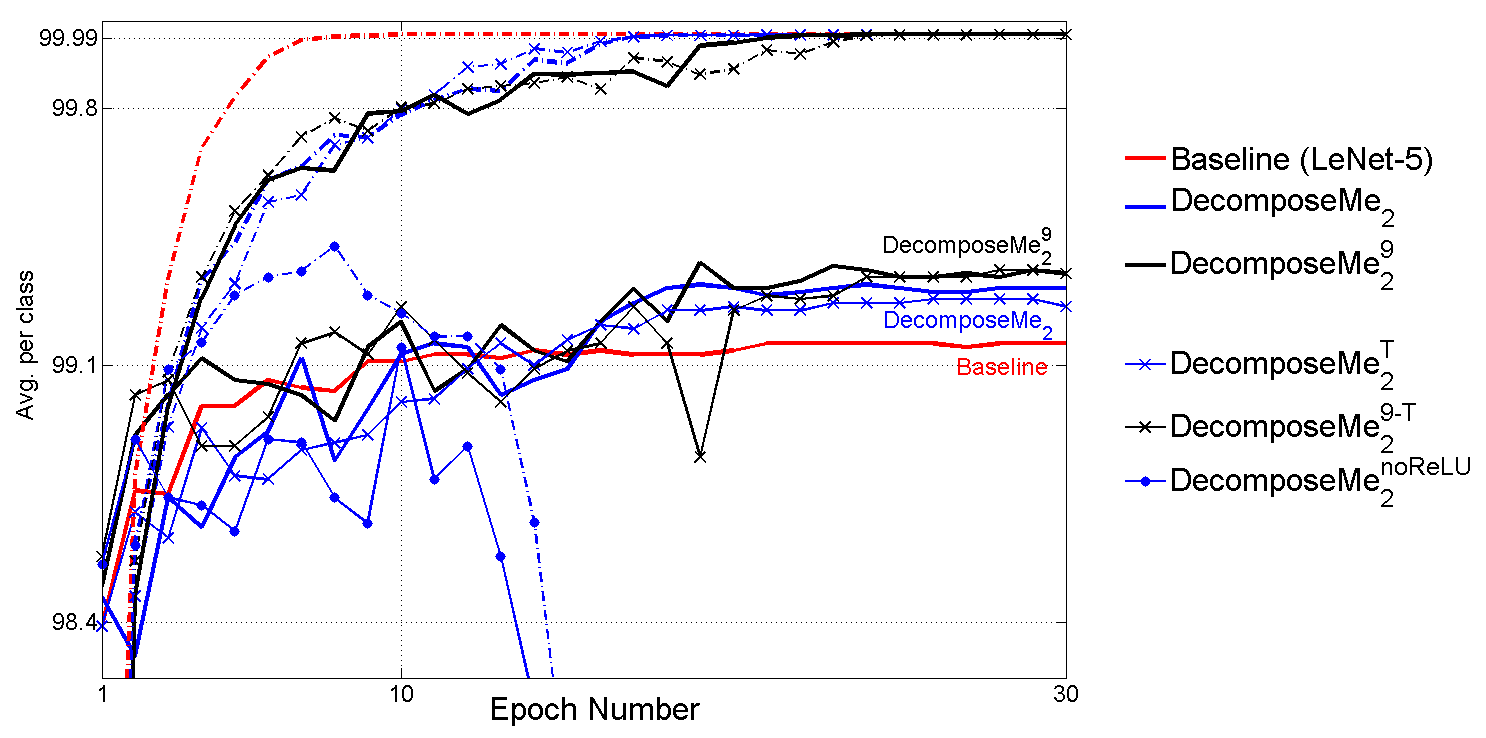}
\end{minipage}&\hspace{-0.35cm}
\begin{minipage}[c]{0.5\textwidth}
\centering\scriptsize
\begin{tabular}{llccc}
&&Avg. (\%) & \multicolumn{1}{c}{\#Params\footnote{\label{ft:p}\scriptsize{Total number of parameters.}}}& \multicolumn{1}{c}{$\max{d}$\footnote{\label{d}\scriptsize{$\max d$ is the largest kernel size in the network.}}}\\
\hline
\hspace{-0.25cm}\multirow{7}{*}{\begin{sideways}LeNet\end{sideways}}&\hspace{-0.05cm}Baseline (retrain)&99.1&52.0K&5\\
\cline{2-5}\vspace{-0.25cm}\\
&\hspace{-0.05cm}DecomposeMe$_1^{tan}$&99.2&53.8K&5\\
&\hspace{-0.05cm}DecomposeMe$_2^{tan}$&99.2&25.6K&5\\
&\hspace{-0.05cm}DecomposeMe$_2$&99.3&22.0K&5\\[0.75ex]
&\hspace{-0.05cm}LeNet$^{9}$&99.2&53.8K&9\\
&\hspace{-0.05cm}DecomposeMe$_2^9$&99.4&22.0K&9\\
\hline
\end{tabular}
\end{minipage}\\
(a)&(b)\\
\end{tabular}
\vspace{-0.4cm}\caption{\textbf{MNIST.} (a) Training curves for
different instances of our proposal and the baseline. Bold lines
represent test accuracy, and dashed ones training accuracy.
${noReLU}$ stands for layers without non-linearity in-between
convolutions. b) Summary of average per class accuracy and number
of parameters for different instances of our architecture together
with the baseline.}\label{fig:MNIST}\vspace{-0.55cm}
\end{figure}

\subsubsection{MNIST and CIFAR-10}
\vspace{-0.15cm}
As a sanity check, we first run experiments on the MNIST and CIFAR-10 datasets.\\
\textbf{MNIST}~\cite{MNIST} is a database of handwritten digits, consisting of a training set of $60.000$ images and a test set of $10.000$ images. All digits in the database have been size-normalized and centered in a fixed-size image. For this experiment we consider the LeNet model proposed in~\cite{LeNet} consisting of two convolutional layers with $5\times 5$ kernels, each one followed by max-pooling layers and hyperbolic tangents as non-linear layers, and two fully connected layers. We first gradually substitute convolutional layers for decomposed layers maintaining the number of output filters (referred to as DecomposeMe$_i^{tan}$ since this model keeps the hyperbolic tangent between convolutional layers). Then, we conduct an additional experiment setting the non-linearities between the convolutional layers to rectified linear units. In this case, we also consider a larger kernel size of $9$ referred to as DecomposeMe$_2^9$.

Figure~\ref{fig:MNIST} summarizes the results for the baseline together with four instances of our proposal. As shown, decompositions systematically outperform the baseline and, when multiple layers are decomposed, significantly reduce the number of parameters in the network. In addition, the performance improves for larger kernel sizes in the first layer. Performance curves for these and additional instances with different filter compositions or excluding the non-linearity in-between decomposed layers are shown in~\fig{fig:MNIST}a. As shown by DecomposeMe$_i^{noReLU}$, adding the non-linearity is necessary. Looking at the graph one can see that the structure without non-linearity learns adequately at the beginning and then performance drops drastically after a few iterations. Large scale experiments presented in the next section will also confirm the need of a non-linearity in between 1D convolutions. More importantly, as a consequence of the reduced number of parameters, the gap between training and testing accuracy decreases when using decomposed layers which indicates that the structures  are less prone to overfitting. This is evident for instance in the $10$th epoch where our proposed method provides similar test accuracy to the baseline, however, the training data accuracy is significantly larger for the baseline, see Figure~\ref{fig:MNIST}.

\textbf{CIFAR-10}~\cite{CIFAR-10} is a database consisting of $50.000$ training and $10.000$ testing RGB images with a resolution of $32\times 32$ pixels split into $10$ classes. We consider the CIFAR-10 quick model consisting of $5$ convolutional layers with kernels of size $5\times 5$~\cite{quick}.

\begin{figure}[!t]
\centering
\begin{tabular}{cc}
\begin{minipage}[c]{0.58\textwidth}
\centering
\hspace{-0.35cm}\includegraphics[width=0.7\columnwidth]{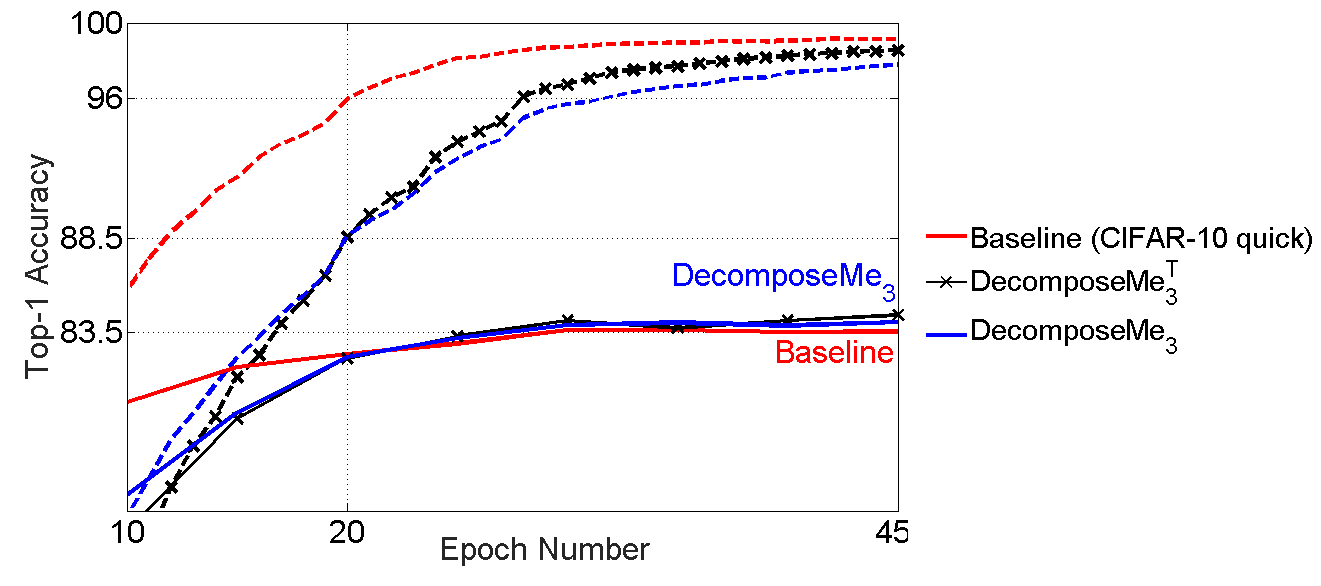}
\end{minipage}&\hspace{-0.35cm}
\begin{minipage}[c]{0.4\textwidth}
\centering\scriptsize
\begin{tabular}{llccc}
&&Top-1 & \multicolumn{1}{c}{\#Params}& \multicolumn{1}{c}{$\max{d}$}\\
 \hline
 \hspace{-0.25cm}\multirow{6}{*}{\begin{sideways}CIFAR-10 quick\end{sideways}}&\hspace{-0.05cm}Baseline (retrained)& 83.8& 5M &5\\
 \cline{2-5}\\
 &\hspace{-0.05cm}DecomposeMe$_1$&84.2&4.7M&5\\
 &\hspace{-0.05cm}DecomposeMe$_3^s$&84.4&3.6M&5\\
 &\hspace{-0.05cm}DecomposeMe$_3$&83.5&2.4M&5\\
 &\hspace{-0.05cm}DecomposeMe$_3^2$&81.2&800K&5\\
 \hline
 \end{tabular}
\end{minipage}\\
(a)&(b)\\
\end{tabular}
\vspace{-0.5cm}\caption{\textbf{CIFAR-10.} a) Training curves for different instances of our proposal and the baseline. As in~\fig{fig:MNIST}, bold lines represent test accuracy, dashed ones training accuracy and a marker indicates a variation in the composition of the layer.  b) Top-1 accuracy and summary of the architecture for different instances of our proposal together with the baseline. The super-index refers to variations of the parameter $L$, see~\eq{eq:dec}.}\label{fig:CIFAR10}\vspace{-0.5cm}
\end{figure}

Figure~\ref{fig:CIFAR10} summarizes the results for the baseline together with an instance of our structure decomposing one layer followed by three instances decomposing all convolutional layers with different $L$ in-between 1D convolutions. As shown, decomposing a single layer provides a slight increment in performance while maintaining the number of parameters. Decomposing additional layers reduces the number of parameters considerably while there is only a slight drop in performance when the reduction is over $50\%$. We have also experimented with different configurations regarding the decomposition --such as horizontal kernels followed by vertical ones and vice versa, or the combination of both-- to verify that the learning process is able to deal with different types of signals. Figure~\ref{fig:CIFAR10}a shows learning curves for the baseline versus our structure with three decomposed layers varying the filter composition: vertical convolution followed by a horizontal one and vice versa (referred to as Decomposed$_3^T$). As we can see in these plots, decomposed layers provide a smaller gap between training and testing accuracy and thus reduce overfitting while maintaining performance. For instance, after $20$ epochs, all the structures provide the same test performance but the training performance of the baseline is $8\%$ higher. These and additional empirical results (not reported) show that the performance is invariant to permutations of the order of the tensors and that there is no significant benefit in combining two types of configurations. Similarly, we have also experimented with substituting the basic architecture for a Network in Network~\cite{NiN:LinCY13} implementation which renders similar benefits when only the first layer is decomposed. In that case, our architecture achieves an increment of $1\%$ in performance.
\subsubsection{Large-Scale Experiments: ImageNet and Places2}
\textbf{\\Datasets.} We now focus on two large-scale datasets: ImageNet~\cite{DengDSLL009} and Places2~\cite{Places2}. ImageNet is a large-scale dataset with over $15$ million labelled images split into $22.000$ categories. We used the ILSVRC-2012~\cite{ImageNet} subset of images consisting of 1.2 million images for training and $50.000$ images for validation. Places2~\cite{Places2} is a large-scale dataset created specifically for training systems targeting high-level visual understanding~\cite{Places2} tasks. This dataset consists of more than $10$ million training images with $401$ unique scene categories and 20000 images for validation. The database comprises between 5000 and 30000 training images per category which is consistent with real-world frequencies of occurrence.

\textbf{Deep Models.} We consider two network structures: the AlexNetOWTBn in~\cite{Krizhevsky14} and the B-net in~\cite{Simonyan14c}(VGG-B). AlexNetOWTBn is the "one weird trick" variation (OWT) of AlexNet~\cite{Krizhevsky_imagenetclassification} where we adopt batch normalization (Bn) after each convolutional layer~\cite{IoffeS15}. B-net~\cite{Simonyan14c} is the B version of the VGG structure and consists of $10$ convolutional layers with max-pooling every two of these convolutions. We consider decompositions in each of those layers, reducing the number of kernels where appropriate. For B-net models, we consider two types of weight initialization: Xavier~\cite{GlorotAISTATS2010} (referred to as DecomposeMe$_i^X$) and Kaiming~\cite{KaimingICCV2015} which we adopt as default configuration since we obtained slightly better results in this case. In both cases, bias terms were set to $0$. Models were trained for a total of $55$ epochs with $10000$ batches per epoch and a batch size of $96$ and $24$ for AlexNetOWTBn and B-net respectively.

\begin{table}[!t]
\caption{\textbf{ImageNet-Places2:} Summary of top-1 accuracy on the validation set for instances of our proposal and baselines on (a) ImageNet and (b) Places2. ConvP stands for the number of parameters in 2D convolutional layers. FCP stands for the number of parameters in fully connected layers and $\max d$ refers to the largest kernel size in the network.}\label{tab:ImageNet-Places}
\centering \scriptsize
\vspace{-0.3cm}
\hspace{0.35cm}\begin{tabular}{cc}
\begin{tabular}{L{0.0005cm}L{0.5cm}lccc}
&&&\hspace{-0.15cm}Top-1 & \multicolumn{1}{c}{\#ConvP}  & \multicolumn{1}{c}{\#FCP}\\
\hline
\hspace{-0.25cm}\multirow{13}{*}{\begin{sideways}AlexNetOWTBn\end{sideways}}&\hspace{-0.10cm}\multirow{11}{*}{\begin{sideways}Full\end{sideways}}&\hspace{-0.35cm}MatConvNet~\cite{VedaldiL14}&\hspace{-0.15cm}57.9 & 2.47M&58.6M\\
&&\hspace{-0.35cm}Baseline (retrain)&\hspace{-0.15cm}57.1 & 2.47M&58.6M\\
\cline{3-6}
&&\hspace{-0.35cm}DecomposeMe$_1$&\hspace{-0.15cm}59.0 & 2.47M &58.6M\\
&&\hspace{-0.35cm}DecomposeMe$_1^{xl}$&\hspace{-0.15cm}59.1  & 2.47M &58.6M\\[0.7ex]
&&\hspace{-0.35cm}DecomposeMe$_1^T$&\hspace{-0.15cm}61.1 & 2.30M&58.6M\\[0.9ex]
&&\hspace{-0.35cm}DecomposeMe$_2$&\hspace{-0.15cm}61.3 & 2.32M&58.6M \\[0.9ex]
&&\hspace{-0.35cm}DecomposeMe$_3$&\hspace{-0.15cm}61.8  & 2.10M &58.6M\\[0.7ex]
&&\hspace{-0.35cm}DecomposeMe$_3^{xl}$&\hspace{-0.15cm}59.4 & 933K &58.6M\\
\cline{2-6}
&\hspace{-0.10cm}\multirow{4}{*}{\begin{sideways}Compact\end{sideways}}&\hspace{-0.35cm}AlexNetOWTBn$^C$&\hspace{-0.15cm}54.7& 2.47M&9.2M\\[0.9ex]
&&\hspace{-0.35cm}DecomposeMe$_3^C$&\hspace{-0.15cm}61.3 & 2.10M&9.2M\\
&&\hspace{-0.35cm}DecomposeMe$_4^C$&\hspace{-0.15cm}57.8 & 1.12M&9.2M\\[0.8ex]
\hline
\hline
\hspace{-0.25cm}\multirow{6}{*}{\begin{sideways}B-net\end{sideways}}&\hspace{-0.10cm}\multirow{3}{*}{\begin{sideways}Full\end{sideways}}&\hspace{-0.35cm}Baseline (retrain)&\hspace{-0.15cm}62.5&9.4M&123.5M\\
\cline{3-6}
&&\hspace{-0.35cm}DecomposeMe$_5^X$&\hspace{-0.15cm}57.5&2.4M&123.5M\\[0.6ex]
&&\hspace{-0.35cm}DecomposeMe$_5$&\hspace{-0.15cm}57.8&2.4M&123.5M\\[0.6ex]
\cline{2-6}
&\hspace{-0.10cm}\multirow{4}{*}{\begin{sideways}Compact\end{sideways}}&\hspace{-0.35cm}B-Net$^C$&\hspace{-0.15cm}61.1&9.4M&25.0M\\[0.9ex]
&&\hspace{-0.35cm}DecomposeMe$_5^{C\mbox{-}X}$&\hspace{-0.15cm}56.9&2.4M&25.0M\\[0.6ex]
&&\hspace{-0.35cm}DecomposeMe$_5^{C}$&\hspace{-0.15cm}57.0&2.4M&25.0M\\[0.6ex]
\hline
&&\hspace{-0.35cm}DecomposeMe$_8^{C}$&\hspace{-0.15cm}65.4&7.0M&8.19M\\
&&\hspace{-0.35cm}DecomposeMe$_8^{C-\textnormal{\scriptsize{avg}}}$&\hspace{-0.15cm}66.2&7.0M&512K\\
\hline
\end{tabular}&\hspace{0.5cm}
\begin{tabular}{L{0.0005cm}L{0.5cm}lccc}
&&&\hspace{-0.15cm}Top-1 & \multicolumn{1}{c}{\#ConvP}  & \multicolumn{1}{c}{\#FCP}\\
\hline
\hspace{-0.25cm}\multirow{9}{*}{\begin{sideways}AlexNetOWTBn\end{sideways}}&\hspace{-0.10cm}\multirow{4}{*}{\begin{sideways}Full\end{sideways}}&\hspace{-0.35cm}Baseline (retrain)&\hspace{-0.15cm}44.5 & 2.47M&56.1M\\[1ex]
\cline{3-6}\\
&&\hspace{-0.35cm}DecomposeMe$_5$&\hspace{-0.15cm}45.2 & 1.52M &56.1M\\[1ex]
\cline{2-6}\\
&\hspace{-0.10cm}\multirow{4}{*}{\begin{sideways}Compact\end{sideways}}&\hspace{-0.35cm}AlexNetOWTBn$^C$&\hspace{-0.15cm}41.1 & 2.47M&3.7M\\[1ex]
\\
&&\hspace{-0.35cm}DecomposeMe$_3^C$&\hspace{-0.15cm}43.5 & 2.10M&3.7M\\[1ex]
\hline
\hline
\hspace{-0.25cm}\multirow{10}{*}{\begin{sideways}B-net\end{sideways}}&\hspace{-0.10cm}\multirow{4}{*}{\begin{sideways}Full\end{sideways}}&\hspace{-0.35cm}Baseline (retrain)&\hspace{-0.15cm}44.0&9.4M&121M\\[0.8ex]
\cline{3-6}
&&\hspace{-0.35cm}DecomposeMe$_6$&\hspace{-0.15cm}43.8&3.0M&121M\\[0.8ex]
&&\hspace{-0.35cm}DecomposeMe$_6^B$&\hspace{-0.15cm}43.6&4.3M&121M\\[0.8ex]
\cline{2-6}\\
&\hspace{-0.10cm}\multirow{3}{*}{\begin{sideways}Compact\end{sideways}}&\hspace{-0.35cm}B-Net$^C$&\hspace{-0.15cm}43.1&9.4M&10M\\[0.8ex]
\\
&&\hspace{-0.35cm}DecomposeMe$_6^C$&\hspace{-0.15cm}43.8&3.1M&10M\\[0.8ex]
&&\hspace{-0.35cm}DecomposeMe$_5^C$&\hspace{-0.15cm}41.3&2.7M&10M\\[0.8ex]
\hline
&&\hspace{-0.35cm}DecomposeMe$_8^{C-256}$&\hspace{-0.15cm}47.4&7.0M&3.2M\\[0.8ex]
\hline
\end{tabular}\\(a)&(b)\\
\end{tabular}
\vspace{-0.6cm}
\end{table}
\textbf{Network Analysis.} We analyze several modifications of the models to better understand the contribution of the proposed approach. First, we study the effect of including non-linearities in-between convolutional layers and different types of filter compositions such as horizontal kernels followed by vertical ones and vice versa, or the combination of both. Second, following the trend of recent architectures~\cite{GoogLeNet_CVPR2015,ResNet:2015,simplicity:2015} we remove intermediate fully connected layers of the models to compare the performance of the convolutional layers. These compact models solely include a fully connected layer to produce the desired number of outputs ($1000$ and $401$ neurons in ImageNet and Places2 respectively). Figure~\ref{fig:DecMEInstances}b shows a comparison between original and compact models. As a direct consequence of removing fully connected layers, the number of parameters drops drastically. For comprehensive comparison we also train and report results for the baselines models in their compact form. Compact models do not use DropOut~\cite{DropOut:2012}.

\begin{figure*}[!t]
\begin{center}
\begin{tabular}{c|c}
\hspace{-0.05cm}\includegraphics[width=0.45\columnwidth,height=2.5cm]{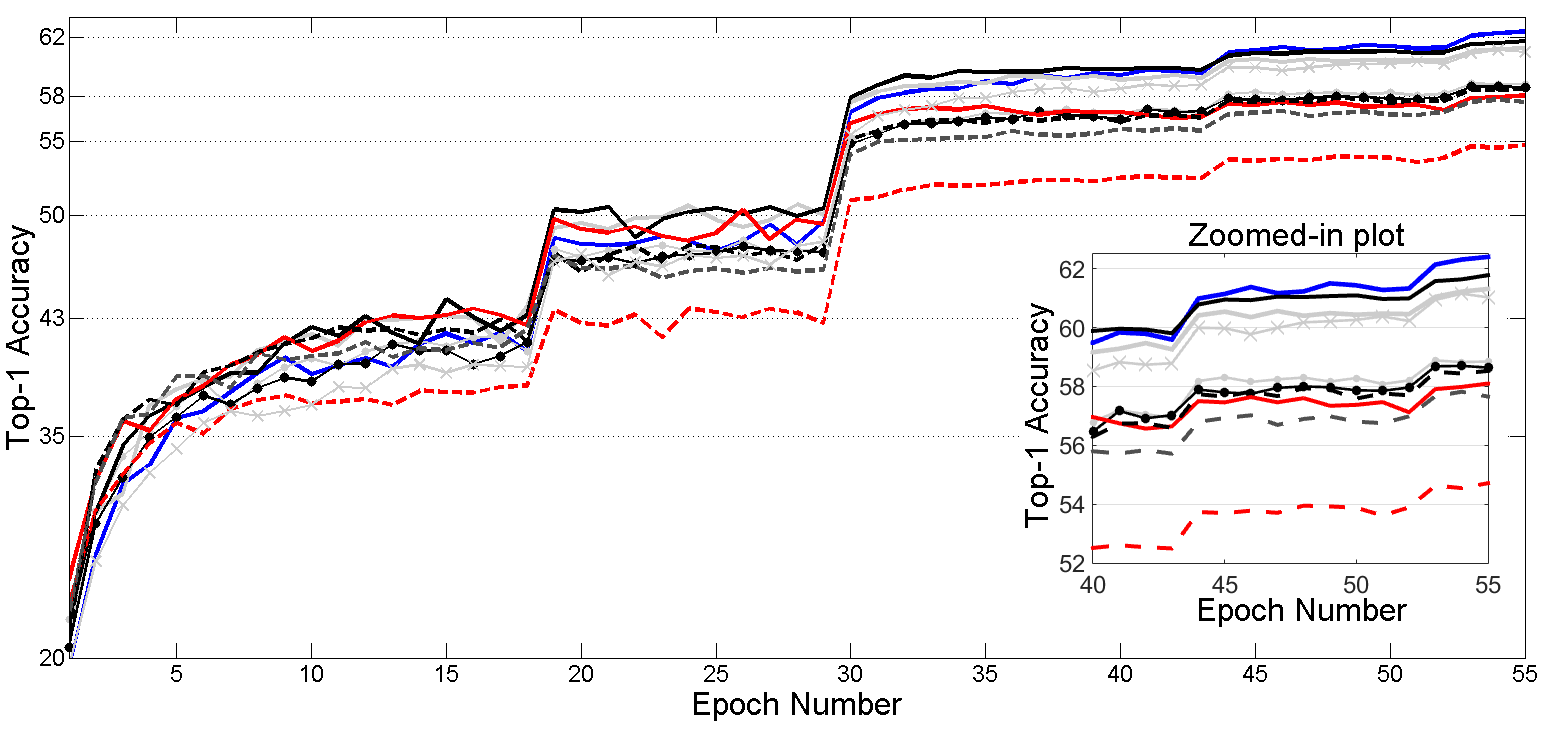}&\includegraphics[width=0.45\columnwidth,height=2.5cm]{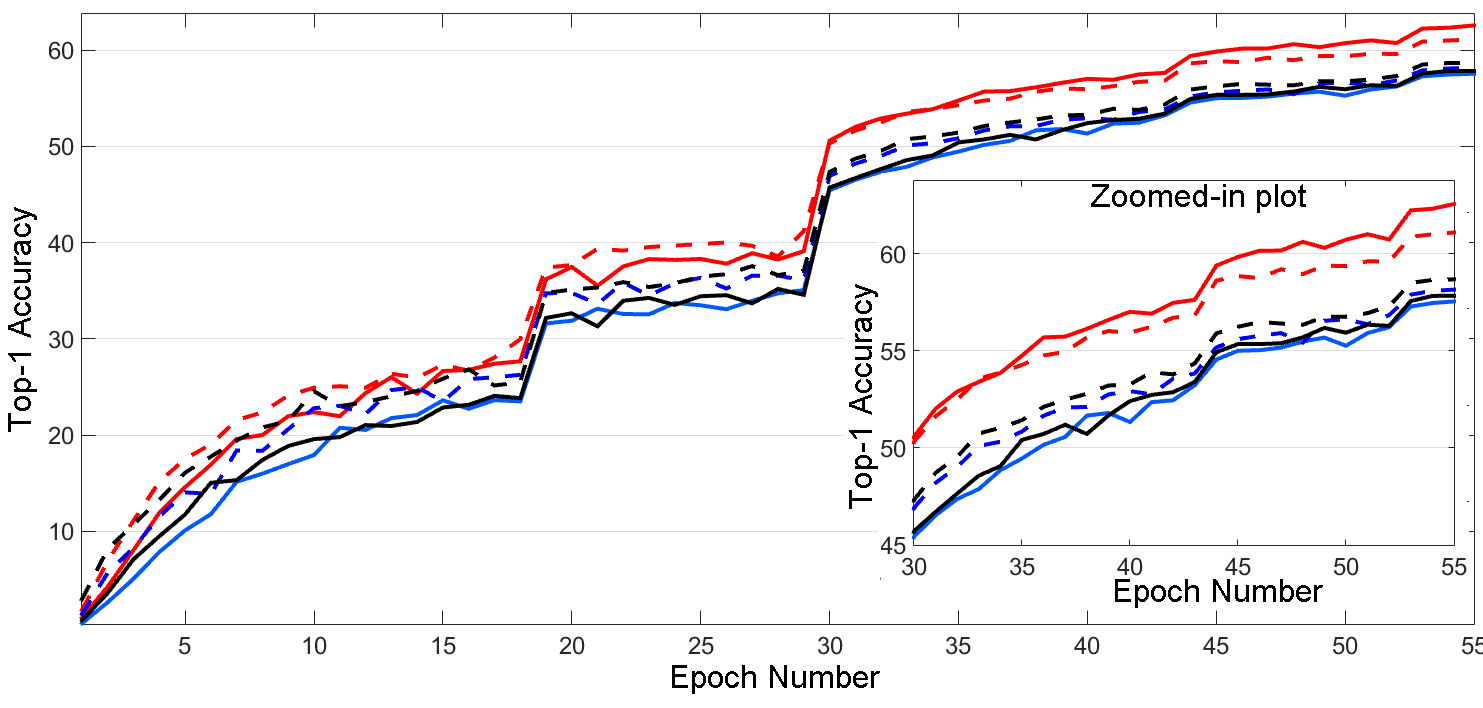}\\
(a)&(b)\\
\hspace{-0.05cm}\includegraphics[width=0.48\columnwidth,height=2.05cm]{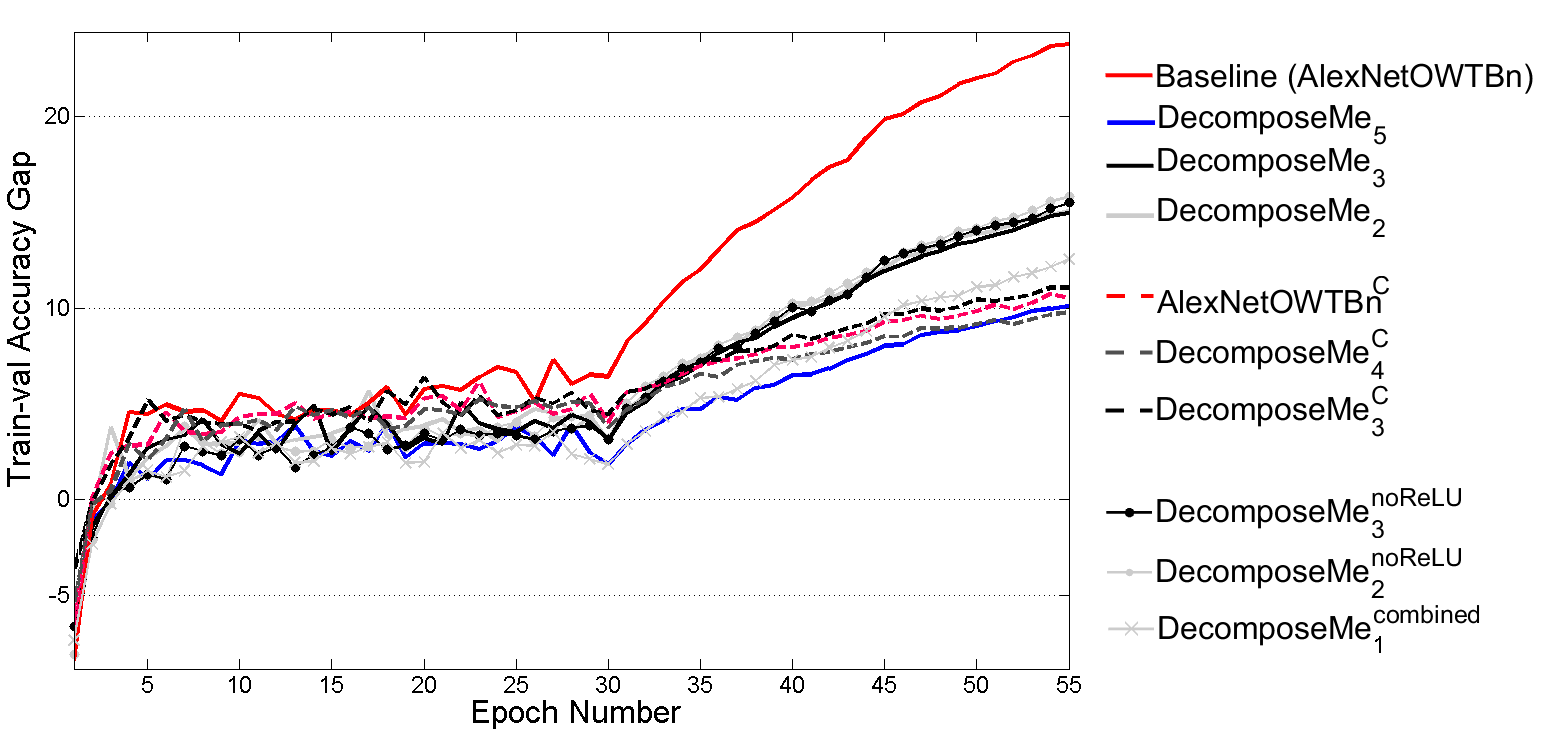}&\includegraphics[width=0.45\columnwidth,height=2.05cm]{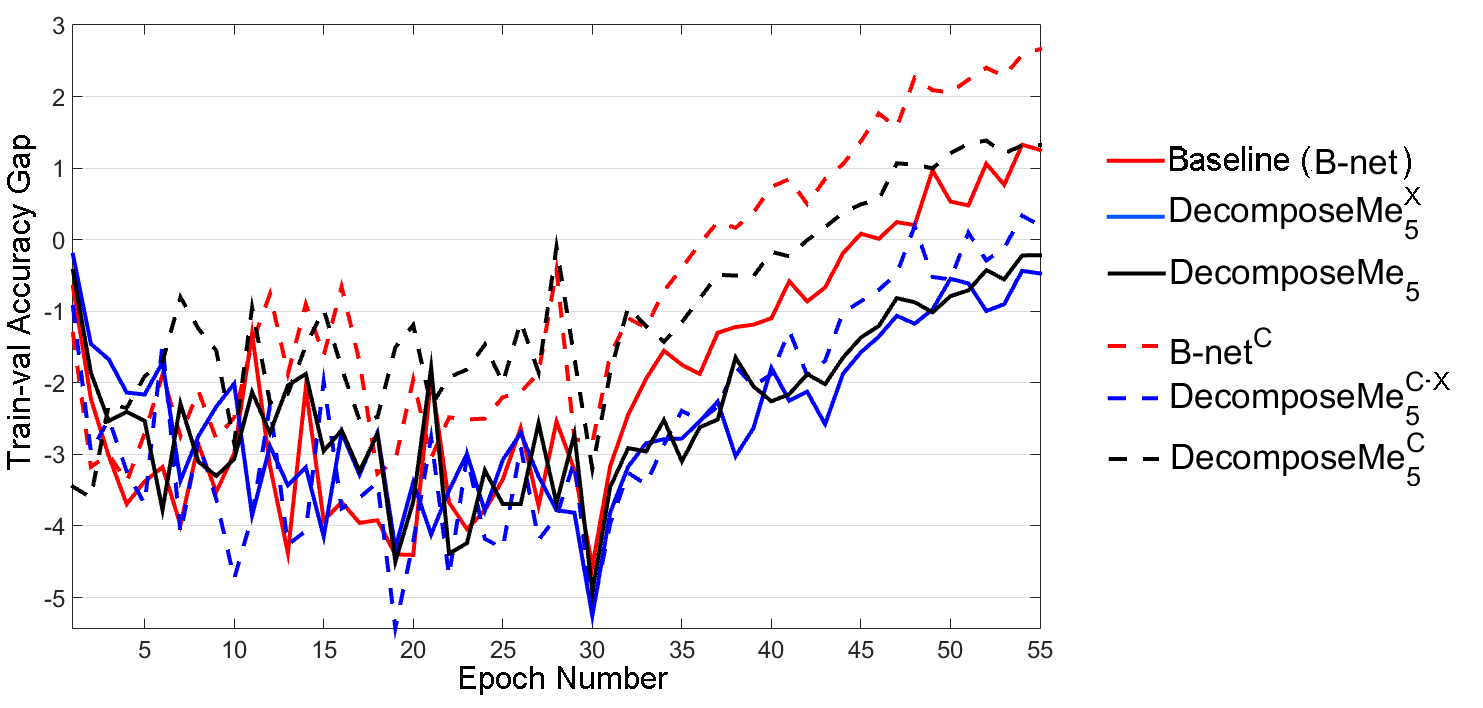}\\
(c)&(d)\\
\end{tabular}
\end{center}
\vspace{-0.55cm}\caption{\textbf{ImageNet.} Training curves for
representative instances of our proposal and the baselines. (a)
and (c) show top-1 accuracy on the validation set and train-val
gap plots respectively for instances of AlexNetOWTBn. (b) and (d)
show the corresponding curves for instances of
B-net.}\label{fig:ImageNetTraining} \vspace{-0.6cm}
\end{figure*}
\textbf{Evaluation.} We measure classification performance as the top-1 accuracy on the validation set using the center crop, named \textbf{Top-1}. We also provide training-validation accuracy gap plots, named \textbf{Train-val gap}. This plot demonstrates the evolution of the difference between train and validation accuracy as the training proceeds~\cite{Batra:ICLR2016}. Overfit models tend to produce a high (positive) gap while underfit models tend to have a similar performance and, therefore, produce a low train validation accuracy gap.

\begin{figure*}[!t]
\begin{center}
\begin{tabular}{c|c}
\hspace{-0.05cm}\includegraphics[width=0.48\columnwidth,height=2.6cm]{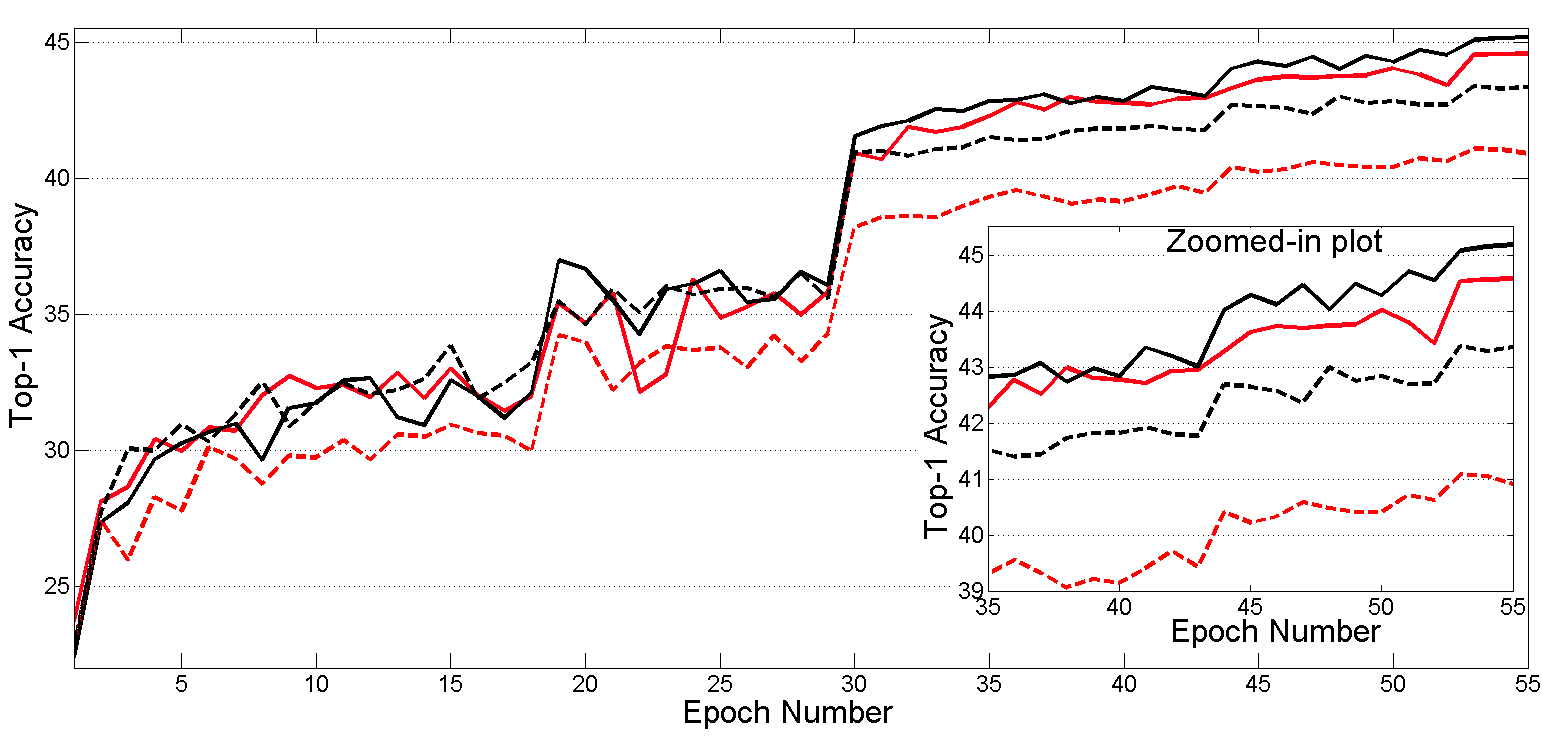}&\includegraphics[width=0.48\columnwidth,height=2.6cm]{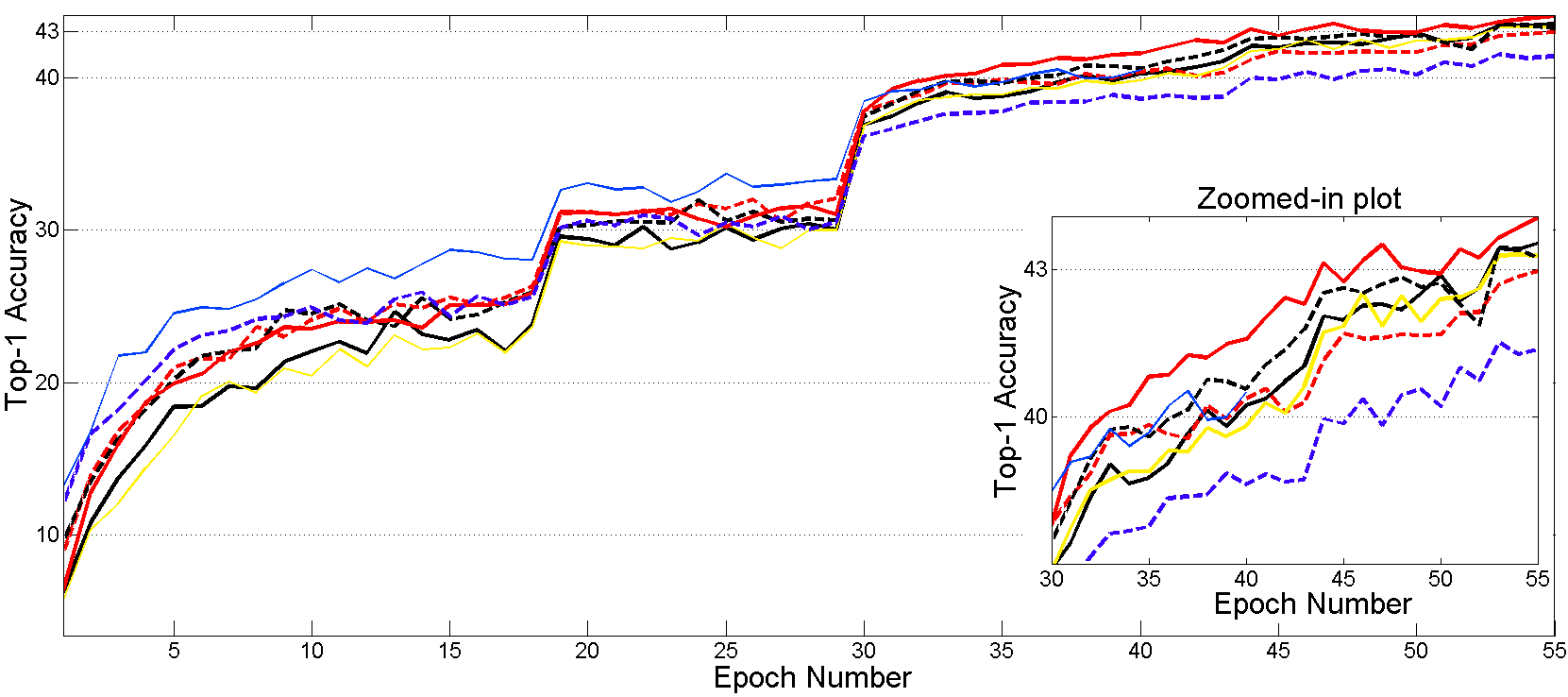}\\
(a)&(b)\\
\hspace{-0.05cm}\includegraphics[width=0.45\columnwidth,height=2.05cm]{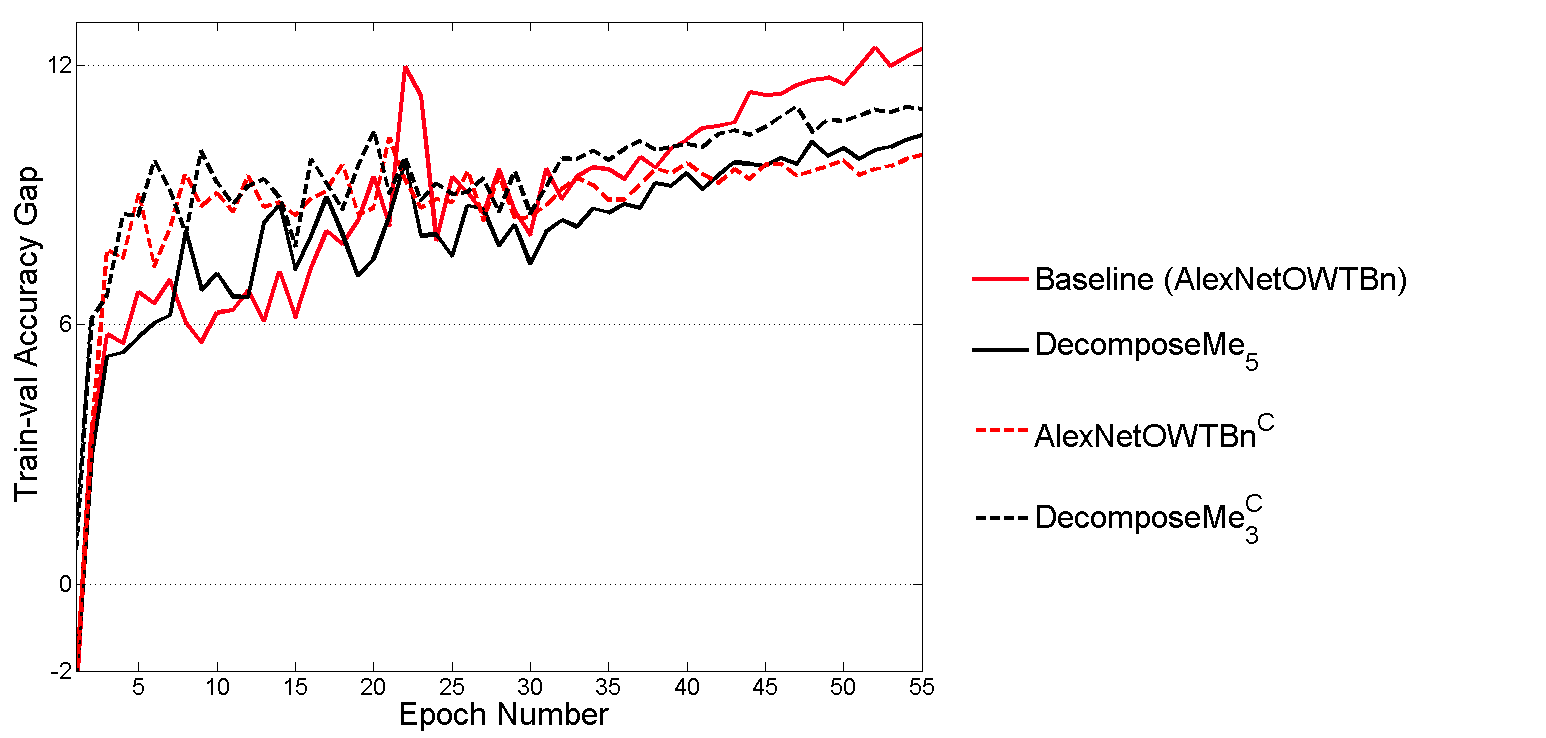}&\hspace{-0.05cm}\includegraphics[width=0.45\columnwidth,height=2.05cm]{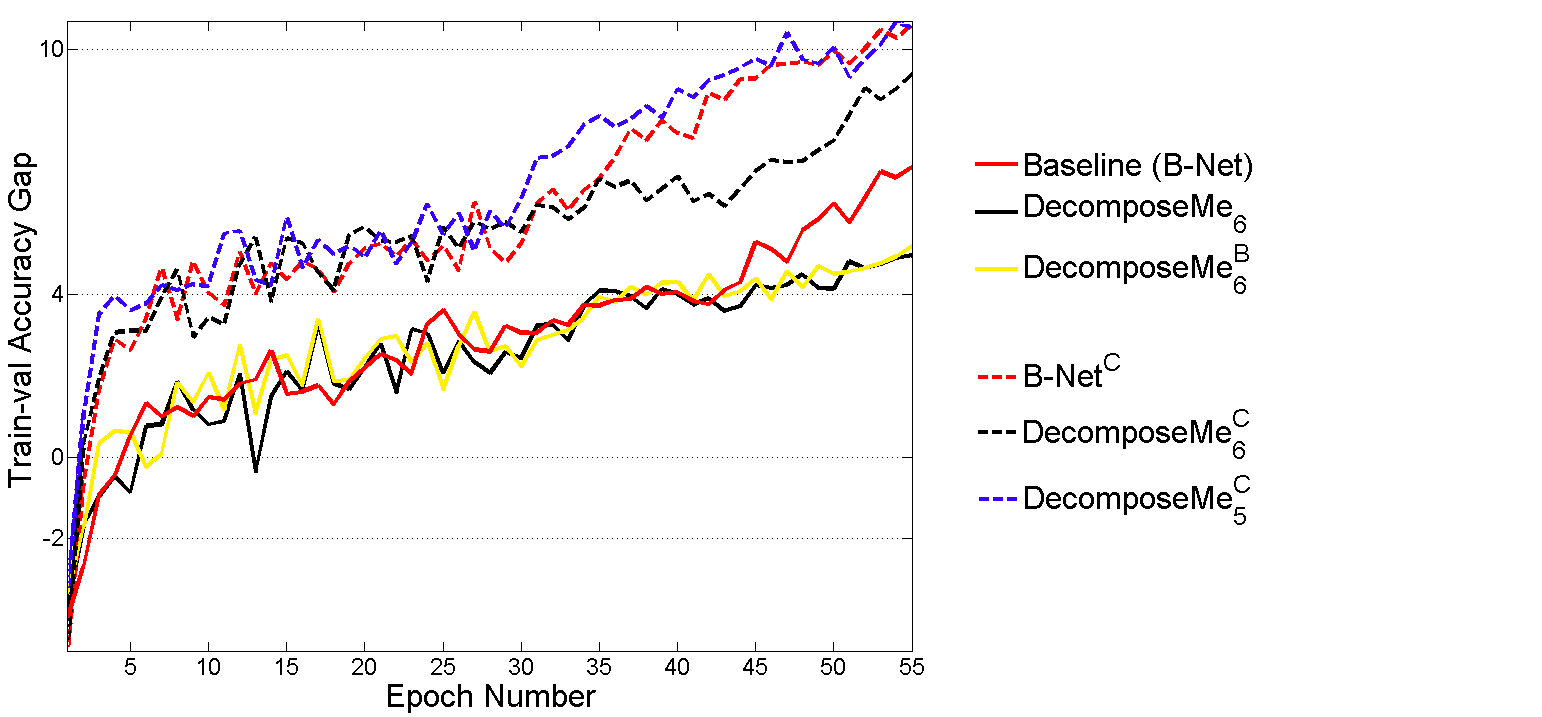}\\
(c)&(d)\\
\end{tabular}
\end{center}
\vspace{-0.5cm}
\caption{\textbf{Places2.} Training curves for representative instances of our proposal and the baselines. (a) and (c) show top-1 accuracy on the validation set and train-val gap plots respectively for instances of AlexNetOWTBn. (b) and (d) show the corresponding curves for instances of B-net.}\label{fig:Places2Training}
\vspace{-0.5cm}
\end{figure*}

\textbf{Experimental results.}
A summary of the results is listed in~\tab{tab:ImageNet-Places}a and~\tab{tab:ImageNet-Places}b for ImageNet and Places2 respectively. Training plots for selected instances of AlexNetOWTBn and B-Net are shown in~\fig{fig:ImageNetTraining} and in~\fig{fig:Places2Training} for ImageNet and Places2 respectively. As shown in~\tab{tab:ImageNet-Places}a, for AlexNetOWTBn, the number of parameters is only reduced when more than one layer is decomposed. This is expected since, in the AlexNetOWTBn structure, each decomposed layer introduces an additional convolutional layer (see the complexity analysis in~\sect{sect:compcost}). However, despite the slightly larger number of parameters, there is an increment in performance. Empirically, we find similar results when a single layer on OverFeat~\cite{sermanet-iclr-14} is decomposed. In that case, there is a performance increment of $2\%$ with respect to the baseline. This suggests that simplified kernels (compositions of 1D kernels) actually help during the training process and the effective capacity of the models increases with the additional non-linear layers.

More substantial changes occur when additional layers are decomposed. The network is then able to produce better results and at the same time reduce the amount of parameters being used. The reduction in the number of parameters is even more substantial when the third layer is decomposed. In this case, the model is still able to perform better than the baseline using only $37.5\%$ of the parameters with respect to the baseline. These results suggest that simplifying ConvNets using the proposed decomposition method not only reduces the amount of parameters required but also outperforms equivalent models learning the complete filter. As in the MNIST experiments, we see no significant difference between variations in the composition of the filters such as horizontal kernels followed by vertical ones (referred to as DecomposeMe$_i^T$). Therefore, we select vertical kernels followed by horizontal ones as the default choice which leads to computational benefits due to memory alignment.

Training curves comparing the effect of including non-linearities in-between decomposed layers are shown in Figure~\ref{fig:ImageNetTraining}a (referred to as DecomposeMe$_2^{noReLU}$ and DecomposeMe$_3^{noReLU}$). As shown, models including non-linearities outperform their equivalent not using rectified kernels independently of the number of decomposed layers. These results suggests that the additional non-linearity in-between each decomposed layer increases the effective capacity of the structure.
Interestingly, we can also see in sub figures $c$ and $d$ of Figure~\ref{fig:ImageNetTraining} and Figure~\ref{fig:Places2Training} that decomposed layers consistently produce training curves with a smaller gap between training and validation accuracy. From these results, we can infer that low-rank filters help in the regularization process during training. These results are in line with the conclusions drawn in~\cite{NIPS+Fergus:2014}. From these results, we can conclude that our proposed method is less prone to overfitting measured as the gap between training and validation accuracy.

We now focus on results obtained using compact networks (referred to using $^C$). First, in Figure~\ref{fig:ImageNetTraining}a and Figure~\ref{fig:Places2Training}a we can see that compact instances of AlexNetOWTBn using decomposed layers outperform their equivalent using fully connected layers. For ImageNet (Figure~\ref{fig:ImageNetTraining}a), compact versions provide competitive results compared to the (full) baseline. Compact B-net models on ImageNet provide slightly lower performance than their equivalent full models as shon in Figure~\ref{fig:ImageNetTraining}b. Nevertheless, the drop of performance is negligible. For these compact models we also observe in Figure~\ref{fig:ImageNetTraining}c that the gap between training and validation accuracy is negative during most of the training process and, therefore, suggests that these models are too small for this particular dataset. The behavior of decomposed versions of the B-net structure on Places2 is different as shown in Figure~\ref{fig:Places2Training}b and Figure~\ref{fig:Places2Training}d. As summarized in Table~\ref{tab:ImageNet-Places}b, all models provide similar performance on this dataset. These results suggest that 2D filters are, in fact, sub-optimal layers that need additional fully connected layers to improve performance. Compared to the baselines, compact models lead to an even more significant reduction in the total number of parameters, see~\tab{tab:ImageNet-Places}. From these results, we can conclude that using 1D convolution layers not only reduces the number of operations and parameters, but also provides competitive (or better) performance compared to state-of-the-art methods.

\begin{figure}[!t]
\centering
\begin{tabular}{cc}
\begin{minipage}[c]{0.58\textwidth}
\hspace{-0.35cm}\includegraphics[width=\textwidth]{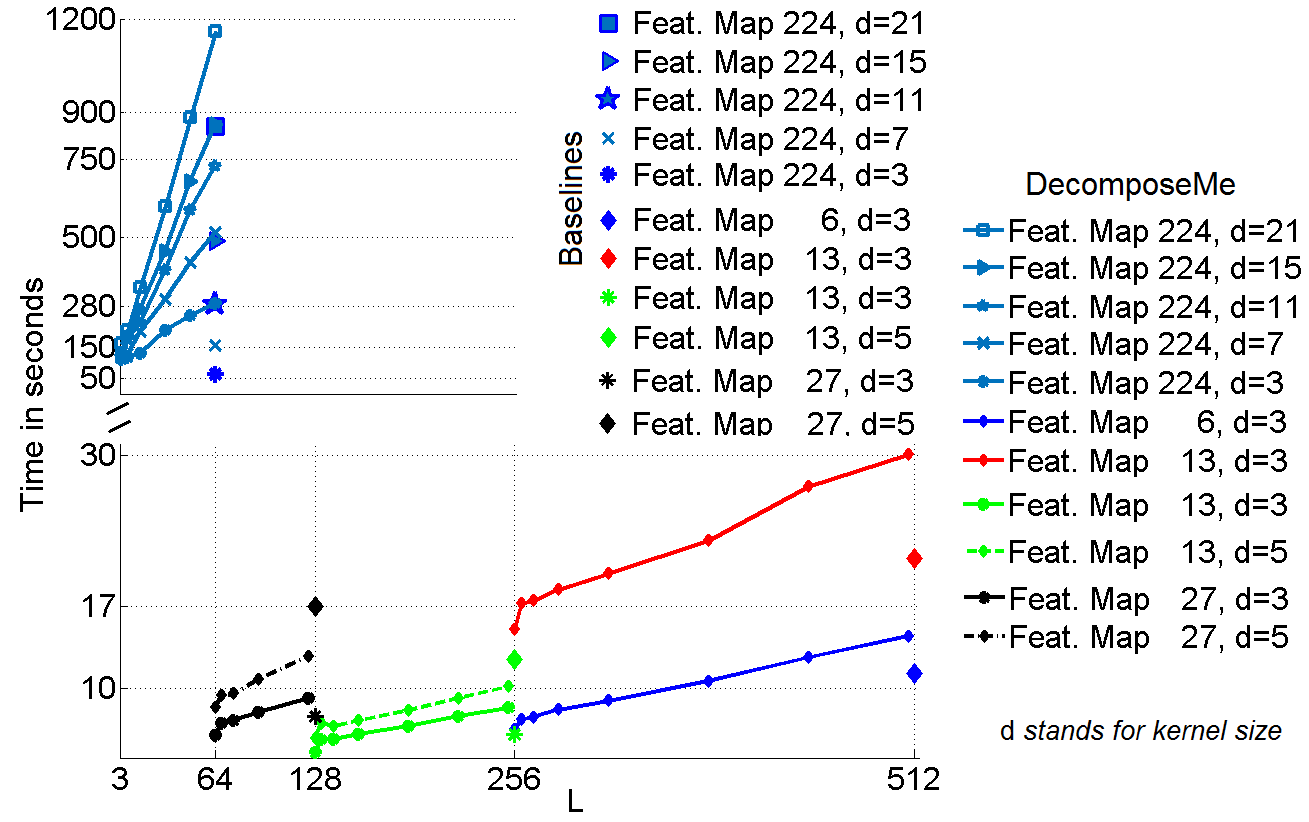}
\end{minipage}&\hspace{-0.1cm}
\begin{minipage}[c]{0.4\textwidth}
\centering\scriptsize
\begin{tabular}{L{0.0005cm}l|cc|}
&Model& Forward Time& Total Time\\
\hline
\hspace{-0.25cm}\multirow{8}{*}{\begin{sideways}AlexNetOWTBn\end{sideways}}&AlexNetOWTBn~\cite{krizhevsky2009learning}&22.45&69.04\\[0.8ex]
\cline{2-4}
&AlexNetOWTBn$^C$&19.98&58.11\\
&DecomposeMe$_3$&28.90&90.17\\
&DecomposeMe$_2$&28.38&88.59\\[1ex]
&DecomposeMe$_3^C$&27.79&83.66\\[0.8ex]
\hline
\hline
\hspace{-0.25cm}\multirow{9}{*}{\begin{sideways}B-Net\end{sideways}}&B-Net~\cite{Places2}&140.45&560.70\\[0.8ex]
\cline{2-4}
&B-Net$^C$&135.13&535.05\\
&DecomposeMe$_6$&56.19&271.52\\
&DecomposeMe$_5$&51.87&252.50\\[0.8ex]
&DecomposeMe$_6^C$&63.89&289.20\\
&DecomposeMe$_5^C$&47.02&226.53\\[1ex]
&DecomposeMe$_8^{C-256}$&38.54&130.13\\
\hline
\end{tabular}
\end{minipage}\\
(a)&(b)\\
\end{tabular}
\vspace{-0.35cm}
\caption{\textbf{Time Analysis.} a) Time as a function of $L$, see~\eq{eq:dec}, and the size of the kernels, $d$. Time is measured as the total of a forward-backward pass of a convolutional layer for the baselines and two 1D convolutions and a rectifier linear unit for decomposed layers. Baselines correspond to convolutional layers using the same number of input-output 2D kernels. Feature map sizes vary corresponding to typical sizes in AlexNetOWT and B-Net. Times are obtained using a batch size of $32$. b) Time benchmarks (in seconds) for the baselines and different instances of our proposed method. Timings are obtained using batches of $8$ RGB images of size $224\times 224$. Both timings are obtained on a Tesla K20m GPU.}\label{fig:Timings} \vspace{-0.5cm}
\end{figure}

The significant reduction in the number of parameters and memory footprint has not only benefits at test time. During training, these compact models make a better use of resources available. For instance, it is possible to increase the batch size to improve the estimation of gradients and, therefore, leverage larger amounts of data. The bottom line of Table~\ref{tab:ImageNet-Places}b shows one additional instance of our method with larger number of decompositions trained with a batch size of $256$, referred to as DecomposeMe$_8^{C-256}$. Please, note that this was not feasible using the baselines. As we can see, the number of parameters of this model is significantly lower than the baseline (e.g., 92\% reduction) and, more importantly, there is a significant improvement in accuracy and computational complexity as we will see in Section~\ref{sect:compcost}.

\vspace{-0.25cm}
\subsection{Complexity analysis}\vspace{-0.1cm}
\label{sect:compcost}
Figure~\ref{fig:Timings}a shows the empirical computational costs of 2D convolutional layers (baselines) and decomposed layers for different representative layers. The plot represents the total time required in a forward-backward pass as a function of $L$. For the baseline, we report the time required solely for the convolution while for decomposed layers we report the combination of 1D convolution, non-linear layer, and 1D convolution. As we can see, the first layer does not produce any benefits time-wise. However, the significant reduction in time occurs for subsequent layers especially for using kernel sizes larger than $3\times 3$. As shown, a more substantial reduction is achieved when $L$ is similar to the number of input filters.

Empirical costs for baselines and instances of decompositions used in our experiments are summarized in~\fig{fig:Timings}b. As expected, we can observe that the amount of time spent during fully connected layers is not meaningful compared to the time required by convolutional layers (see the comparison between AlexNetOWTBn and AlexNetOWTBn$^C$). Besides, substantial savings occur for instances of B-Net models where pairs of layers are decomposed and, therefore, maintaining the number of layers.

A fair comparison with existing low-rank approximation methods~\cite{Fergus:NIPS2014,Jaderberg14b} is difficult as they require a fully pre-trained network to initialize their methods and, they need a fine-tuning process to prevent significant drops in performance. Contrary to them, our method is trained directly from data using a standard initialization. Compared to~\cite{Fergus:NIPS2014}, for ImageNet considering AlexNetOWTBn as a similar network architecture (four convolutional layers and three fully connected layers), we obtain an increment in the top-1 performance of $5.87\%$ with a $5.4x$ reduction in the number of weights. Our result is significantly better than the $2.5x$ reduction in the number of weights with an increment in error (top-5) of $0.02\%$ reported in~\cite{Fergus:NIPS2014}. Best results reported in~\cite{Jaderberg14b} are a speedup of $2.5x$ with no loss in accuracy and a $4.5x$ with a drop of $1\%$ in classification accuracy on ICDAR2003. In our case, our best result is on Places2 where we achieve a $3.5x$ speedup in forward time with a $12.5x$ reduction in the number of parameters and an increment in top-1 classification accuracy of $5.7\%$.
\subsection{Stereo Matching}
\label{subsect:stereo}\vspace{-0.1cm}
The purpose of this experiment is to further demonstrate the applicability of our method when converting existing complex architectures. Accomplishing this, we address the problem of computing the disparity for each pixel in an image given a stereo pair of images. In particular, we use the recent method proposed in~\cite{ZbontarL15} where Zbontar \etal propose a ConvNet that matches patches in a stereo pair. The architecture consists of two feature extraction models, one per image and whose output serves as input for learning the matching network~\cite{ZbontarL15}. The entire process is learned in an end-to-end fashion and provides state-of-the-art results on KITTI2012~\cite{Geiger2013IJRR}.

In this experiment, we focus on converting the feature extraction models to decomposed ones. These modules use four consecutive convolutional layers with kernels of size $3\times 3$ with rectified linear units following each layer. Demonstrating the versatility of our architecture we test two different decompositions as outlined in~\fig{fig:DecMEInstances}c. Firstly, we pair every two convolutional layers and transform them into a decomposed one. Secondly, we consider a unique decomposition that compacts the four layers into a decomposed one using larger kernels of size $9\times 9$. Therefore, both decompositions leverage the same neighborhood of size $9\times 9$ in the input feature map which is the equivalent to four consecutive convolutions of $3\times3$ in the original model. Table~\ref{tab:stereoNets} summarizes the results for these models. We show the numbers after retraining the original network (and, therefore, all randomization is equivalent). Table~\ref{tab:stereoNets} includes the run time for the complete process including the matching network as well as the time required to extract features which is the focus of this experiment. As we can see, our approach significantly reduces the time required to extract features from each image. Nevertheless, this has almost no impact in the overall time which is consistent with the original paper~\cite{ZbontarL15}, as the feature part is not responsible for the majority of the computational cost. More importantly, our proposed method achieves almost the same performance with a significant reduction in the number of parameters.

\begin{table}[!t]
\caption{\textbf{KITTI2012.} a) Results for two instances of our method compared to the baseline. Al models, including the baseline, are retrained from scratch. Time is the forward-backward time over a batch of $8$ images measured on a Tesla K20m. b) The leading submission on the KITTI 2012 leader-board as of 1st November 2015.}
\label{tab:stereoNets}\centering\scriptsize
\vspace{-0.25cm}
\begin{minipage}[c]{\columnwidth}
\begin{tabular}{cc}
\hspace{0.05cm}\begin{tabular}{L{2.4cm}ccccc|c}
&\hspace{-0.15cm}Error &\multicolumn{1}{c}{\hspace{-0.15cm} \#FP\footnote{\scriptsize{FP stands for the number of parameters in convolutional layers. \#Lay refers to the number of (1D or 2D) convolution layers. $\max d$ is the largest kernel size in the network. } }}  &\multicolumn{1}{c}{\hspace{0.10cm}\#Lay} &\multicolumn{1}{c}{\hspace{-0.05cm}$\max{d}$ }&\hspace{-0.05cm}Runtime&Feat. Time\\
\hline
\hspace{-0.15cm}Baseline (retrained)~\cite{ZbontarL15}&\hspace{-0.05cm}2.60&\hspace{0.10cm}339K &\hspace{-0.05cm}4&\hspace{-0.05cm} 3&\hspace{-0.05cm}64.5s&\hspace{-0.05cm}776.9s\\[0.9ex]
\hline
\hspace{-0.15cm}DecomposeMe$^a$&\hspace{-0.05cm}2.66&\hspace{-0.05cm}48K&\hspace{0.10cm}4&\hspace{-0.05cm}5&\hspace{-0.05cm}63.1s&\hspace{-0.05cm}312.9s\\[0.9ex]
\hspace{-0.15cm}DecomposeMe$^b$&\hspace{-0.05cm}2.72&\hspace{-0.05cm}32K&\hspace{0.10cm}2&\hspace{-0.05cm}9&\hspace{-0.05cm}63.5s&\hspace{-0.05cm}281.9s\\[0.9ex]
\hline
\end{tabular}&
\hspace{0.40cm}\begin{tabular}{l | c | c | c | c}\centering\scriptsize
\hspace{-0.25cm}{Method} & ON\footnote{\scriptsize{ON (Out-Noc) and OA (Out-All) stand for percentage of erroneous pixels in non-occluded areas and in total respectively. AN (Avg-Noc) and AA (Avg-All) stand for average disparity / end-point error in non-occluded areas and in total respectively.} } & OA & AN&AA\\ \hline
\hspace{-0.25cm}MC-CNN-acrt~\cite{ZbontarL15} & 2.43& 3.63& 0.7 px & 0.9 px\\
\hspace{-0.25cm}\textbf{DecomposeMe$^b$} & 2.48& 3.69& 0.8 px & 0.9 px\\
\hspace{-0.25cm}MC-CNN & 2.61& 3.84& 0.8 px & 1.0 px\\
\end{tabular}\\
(a)&(b)\\
\end{tabular}
\end{minipage}
\vspace{-0.5cm}
\end{table}
Finally, \tab{tab:stereoNets}b summarizes bench-marking results on KITTI dataset~\cite{Geiger2013IJRR}. Our proposal provides similar results compared to the original network using only $24.3\%$ of the parameters in the feature layers. These are relevant results since our proposed method, without a custom design, can reach similar performance compared to a deep model that was carefully engineered. More importantly, this is achieved using only a fraction of the number of parameters.

\vspace{-0.45cm}
\section{Conclusions}
\vspace{-0.2cm}
In this paper we proposed DecomposeMe. A novel and efficient convolutional neural network architecture based on 1D convolutions. Experiments on large-scale image classification show that our approach improves the classification accuracy while significantly reducing the number of parameters and computational cost. For instance, on Places2 and compared to the VGG-B model, our architecture obtains a relative improvement in top-1 classification accuracy of $7.7\%$ using $92\%$ fewer parameters than VGG-B and with a speed up factor in forward-time of $3.5x$. Additional experiments on stereo matching also demonstrate the general applicability of the proposed architecture. 

{\small
\textbf{\newline Acknowledgment}
The authors thank John Taylor for helpful discussions and continuous support through using the CSIRO high-performance computing facilities. The authors also thank NVIDIA for generous hardware donations.
\bibliographystyle{IEEEtran}
\bibliography{egbib}
}

\end{document}